\documentclass[letterpaper, 10 pt, conference]{ieeeconf}  
\usepackage{epsfig}
\usepackage{amssymb}
\usepackage{amstext}
\usepackage{amsmath}
\usepackage{multicol}
\usepackage{pslatex}
\usepackage{booktabs}
\usepackage{algorithm2e}
\usepackage[bottom]{footmisc}
\usepackage{caption}
\usepackage{etoolbox}
\usepackage{xcolor}
\usepackage{hyperref}
\usepackage{multirow}

\hypersetup{
	colorlinks=true,
	citecolor=red,
	linkcolor=black,
	urlcolor=red
}
\makeatletter
\patchcmd{\@makecaption}
{\scshape}
{}
{}
{}
\makeatletter
\patchcmd{\@makecaption}
{\\}
{.\ }
{}
{}
\makeatother

\UseRawInputEncoding
\IEEEoverridecommandlockouts                              

\title{\LARGE \bf
CoPAD : Multi-source Trajectory Fusion and Cooperative Trajectory Prediction with Anchor-oriented Decoder in V2X Scenarios
}

\author{Kangyu Wu$^{1,2}$ , Jiaqi Qiao$^{1,2}$ and Ya Zhang$^{1,2,*}$
\thanks{$^{1}$School of Automation, Southeast University, Nanjing, China}
\thanks{$^{2}$Key Laboratory of Measurement and Control of Complex Systems of Engineering, Ministry of Education, Nanjing, China}
\thanks{$^{*}$Corresponding author: Y. Zhang(yazhang@seu.edu.cn)}
}

\begin{document}


\maketitle
\thispagestyle{empty}
\pagestyle{empty}

\begin{abstract}

Recently, data-driven trajectory prediction methods have achieved remarkable results, significantly advancing the development of autonomous driving. However, the instability of single-vehicle perception introduces certain limitations to trajectory prediction. In this paper, a novel lightweight framework for cooperative trajectory prediction, CoPAD, is proposed. This framework incorporates a fusion module based on the Hungarian algorithm and Kalman filtering, along with the Past Time Attention (PTA) module, mode attention module and anchor-oriented decoder (AoD). It effectively performs early fusion on multi-source trajectory data from vehicles and road infrastructure, enabling the trajectories with high completeness and accuracy. The PTA module can efficiently capture potential interaction information among historical trajectories, and the mode attention module is proposed to enrich the diversity of predictions. Additionally, the decoder based on sparse anchors is designed to generate the final complete trajectories. Extensive experiments show that CoPAD achieves the state-of-the-art performance on the DAIR-V2X-Seq dataset, validating the effectiveness of the model in cooperative trajectory prediction in V2X scenarios.

\end{abstract}

\section{INTRODUCTION}

In recent years, autonomous driving is a hot research and application area, where safety is the most important issue. Accurate prediction of surrounding vehicle trajectories is crucial for autonomous driving safety, significantly impacting subsequent planning and indirectly ensuring compliance with traffic rules, thereby reducing accident risks. Traditional methods, like Kalman Filtering \cite{abbas2020adaptive} and Markov Models \cite{cai2020comprehensive}, provide strong interpretability based on mathematical models and dynamics principles, but highly depend on the precise information of the model and struggle to extract latent motion patterns from historical data, limiting long-term sequence prediction performance. 
\begin{figure}[!h]
	\centering
	{\epsfig{file = 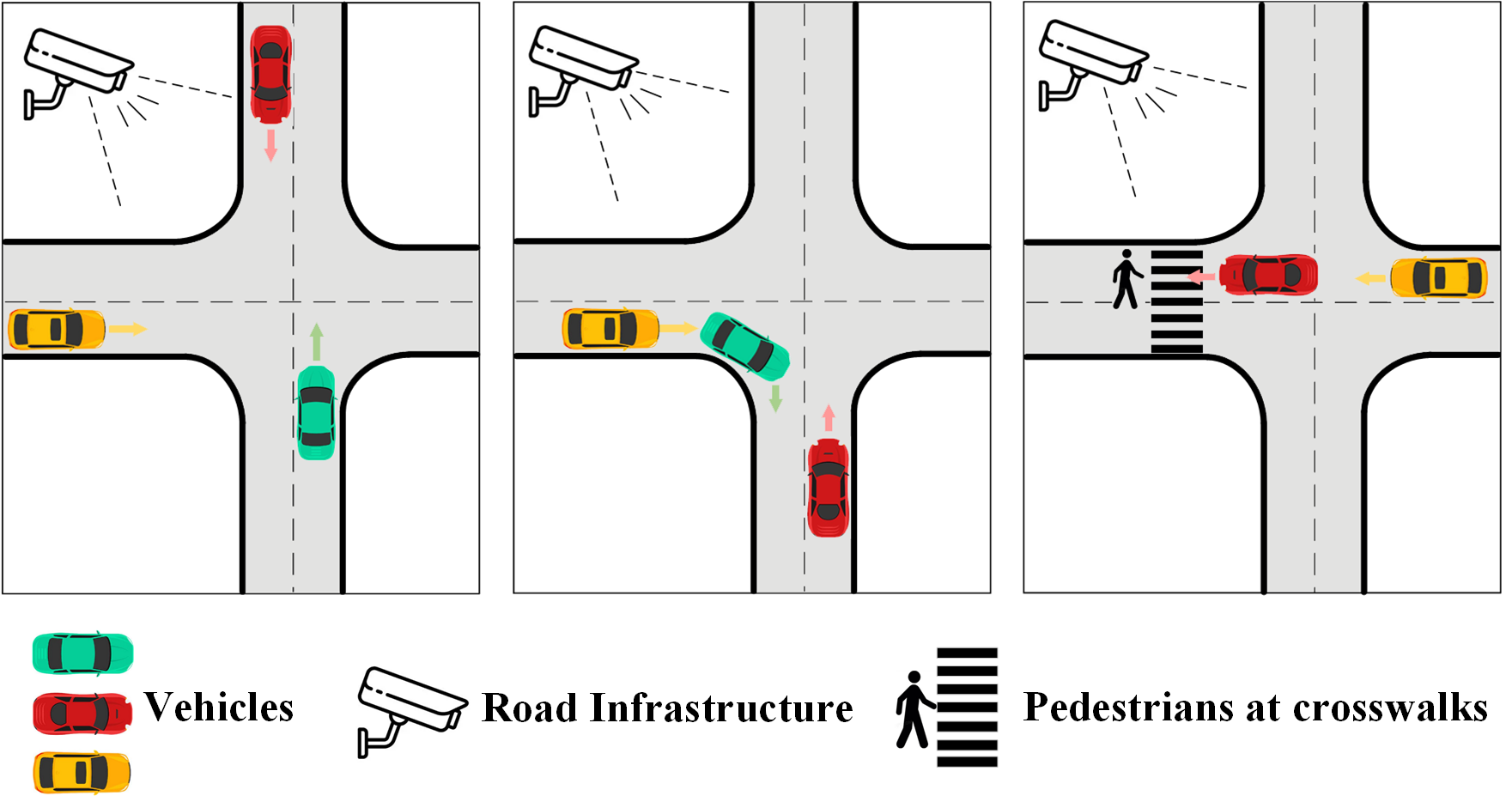, width = 8.8cm}}
	\caption{The schematic diagrams of typical V2X scenarios. Road infrastructure provides an alternative perspective for perception when the vision of the yellow vehicle is limited.}
	\label{fig:1}
\end{figure}

Data-driven deep learning methods are capable of further exploring the deep-level patterns within historical data, enhancing the accuracy of trajectory predictions and, consequently, elevating the safety standards of autonomous driving. Early methods rasterized maps into Bird's-Eye-View (BEV) and employed Convolutional Neural Networks (CNNs) to extract features \cite{chai2019multipath}. However, recent approaches have vectorized high definition maps (HD maps) and agents trajectories \cite{gilles2021home}\cite{gao2020vectornet} representing scenes and motion trajectories by vectors. These innovative encoding methods have reduced computational costs while enhancing the data representation capabilities. The encoded data is processed through models like Recurrent Neural Networks (RNNs), Graph Neural Networks (GNNs), and Transformers, incorporating various attention mechanisms to augment the learning capacity of the models, ultimately outputting multi-modal potential future trajectories \cite{zeng2021lanercnn}\cite{mo2022multi}. Recent research has moved beyond forecasting the future trajectories of a single agent, focusing instead on joint multi-agent prediction \cite{hong2024multi}\cite{zhou2022hivt}. ADAPT \cite{aydemir2023adapt} employs dynamic weight learning to compute the trajectories of all agents within the scene. DCMS \cite{ye2022dcms} frames trajectory prediction as a dynamic problem, accounting for the correlations between consecutive predictions. CMP \cite{kang2024continual} introduces a continuous learning approach that integrates meta-representation learning with an optimal memory buffer retention strategy, effectively addressing the challenge of evolving data streams. CaDeT \cite{pourkeshavarz2024cadet} employs causal disentanglement to isolate the influence of environmental factors on trajectory prediction, enabling the model to automatically adapt to entirely new environments.

The aforementioned data-driven approaches heavily rely on high-quality raw data. However, in real-world scenarios, raw motion trajectories of other vehicles or pedestrians are typically sourced from vehicle-mounted sensors, such as LiDAR or cameras. Due to factors such as excessive distances, or severe occlusion, acquiring complete and accurate data can sometimes be challenging. Consequently, low-quality and sparse data adversely affect trajectory prediction accuracy. Vehicle-to-Everything (V2X) communication \cite{tan2023dynamic} has partially alleviated the instability associated with single-vehicle perception in such environments. Roadside units (RSUs) deployed along the sides or above the road can detect and track agents in real-time, and transmit information to autonomous vehicles. Therefore, communication between road infrastructure and vehicles can effectively address the issue of data sparsity arising from single vehicle perception. In cases where a vehicle's perception data is missing, the infrastructure is capable of providing data from another view, enabling the vehicles to proceed with predictions and decisions normally, which can ensure the safety of vehicles in environments where perception is sparse \cite{su2024makes}, as shown in Fig. \ref{fig:1}.

For the input of multi-source trajectory, we introduce CoPAD, an end-to-end lightweight cooperative trajectory prediction model based on Kalman filtering and GNNs. Multi-source trajectory data will be integrated to compensate for missing data in single vehicle perceptions and enhance the quality of the perceived trajectory, thereby enabling the acquisition of higher-quality prediction results. The fused trajectory, along with the vectorized HD map, is encoded into a graph. The multi-head attention mechanism enables the model to capture the underlying patterns of agent interactions. Subsequently, through the PTA module, the model is able to mine valuable temporal information from past motion scenes and integrate global information across different periods in an interpretable and interactive manner. The mode attention mechanism is incorporated into the model to explore the intrinsic relationships between different modalities. In the final anchor-oriented decoder, we utilize sparse anchors along with Multilayer Perceptrons (MLPs). Experiments have demonstrated that our sparse anchors provide a prior rough location estimate for trajectories to a certain extent. Additionally, compared to other collaborative trajectory prediction models, our model is more lightweight. Feature-level intermediate fusion brings double training parameters, while the early fusion module reduces the overall training parameters subsequently to a certain extent, resulting in faster training and inference speeds for the model.
\begin{figure*}[t]
	\centering
	{\epsfig{file = 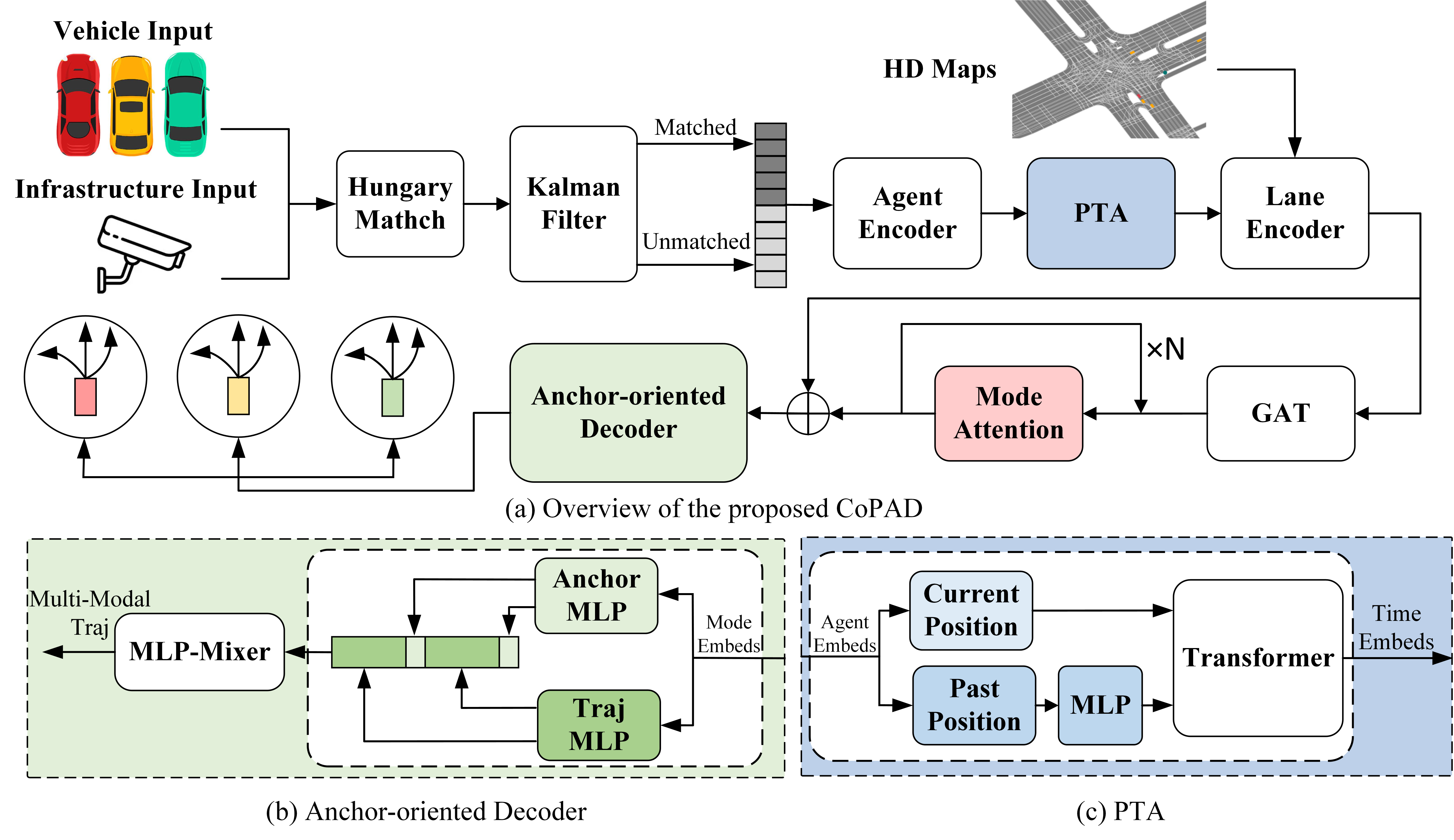, width = \textwidth}}
	\caption{Overview of CoPAD. The input consists of vehicle input $\mathcal{T}_V$, road infrastructure input $\mathcal{T}_I$ and HD maps $\mathcal{M}$. And the output is the multi-modal future trajectories $Y$ of all agents in the scenes. Anchor-oriented Decoder and PTA module are described in detail in figure (b) and (c).}
	\label{fig:2}
\end{figure*}

To evaluate the model, we consider the large-scale dataset DAIR-V2X-Seq, which contains trajectories collected in 60,000 scenarios using vehicle sensors and road infrastructure, encompassing the trajectories of various types of vehicles and pedestrians. However, both types of data sources exhibit some degree of data missingness, accurately reflecting the limitations of single vehicle perception in real-world scenarios.

Overall, the main contributions can be summarized as:

(1)	CoPAD, an end-to-end V2X cooperative trajectory prediction framework is proposed. Based on multi-source trajectory data from DAIR-V2X-Seq, the missing trajectories information is compensated.

(2)	A novel early fusion module based on the Hungarian algorithm and Kalman filtering is designed, capable of accurately matching and fusing multi-source trajectory data, and significantly reducing the training parameters in the collaborative prediction model.

(3)	A global temporal interaction module PTA is proposed, which efficiently interacts with and learns from global information from past scenes at the current moment. Additionally, we introduce an efficient mode attention mechanism that facilitates the exploration of potential connections between different modalities. Finally, the oriented anchors are used in the decoder to locate the trajectory to a certain extent.

\section{RELATED WORK}
\subsection{Cooperative Autonomous Driving}
With the rapid development of technologies such as road infrastructure and information transmission, V2X has become increasingly mature, providing robust support for improving traffic efficiency and safety. A substantial amount of datasets on cooperative autonomous driving has laid the foundation for subsequent research. Some datasets \cite{karvat2024adver} are built upon autonomous driving simulators, while others \cite{hao2024rcooper}\cite{zimmer2024tumtraf}\cite{yu2023v2x}\cite{yu2022dair} are sourced from real-world road traffic scenarios. Most works based on these datasets utilize point clouds as the primary input for perception, often complemented by multi-view image inputs, enabling cooperative tasks such as segmentation \cite{tan2023dynamic}, prediction \cite{yu2023v2x}\cite{wang2024cmp}, detection \cite{hong2024multi}\cite{su2024makes}\cite{zhong2024leveraging}, and planning \cite{glaser2023communication}. Some works \cite{hu2023planning}\cite{chen2024vadv2} have achieved end-to-end autonomous driving solely through image input, yielding astonishing results.


\subsection{GNNs for Motion Forecasting}
Graph Neural Networks \cite{kipf2017semi} inherently emulate the intricate relative interconnections among a collection of entities, making them extensively utilized in modeling traffic scenarios. In such graphs, each node typically represents a pedestrian or a vehicle, while the edges connecting the nodes encode the relative information between them. Through multiple layers of message passing, each node can aggregate and update the features of neighboring nodes. Early methods employed homogeneous GNNs to provide a simple unified representation of both the scenes and the agents \cite{gao2020vectornet}. Subsequent methods shifted towards heterogeneous GNNs \cite{mo2022multi}\cite{ruan2023learning}, enhancing expressive power and interpretability of the models. In terms of model architecture, \cite{gao2020vectornet}\cite{zeng2021lanercnn} utilized graph convolutional networks (GCNs), extending the concept of convolution to graphs. Meanwhile, \cite{mo2022multi}\cite{zhou2022hivt}\cite{ruan2023learning} introduced attention mechanisms into the graph domain to focus on analyzing more important nodes.

\subsection{Attention Mechanism}
Transformer \cite{vaswani2017attention} has brought revolutionary changes to the field of deep learning, with its innovative attention mechanism offering significant improvements for numerous tasks. In trajectory prediction, various models \cite{mo2022multi}\cite{zhou2022hivt}\cite{aydemir2023adapt}\cite{yu2023v2x}\cite{ruan2023learning} utilize attention mechanisms to strengthen the interactions between agents. In this paper, attention mechanisms are employed across multiple dimensions, including time, space, and modality. Inspired by \cite{tang2024hpnet}, we flatten multi-dimensional data to apply attention mechanisms with reduced computational complexity.

\section{METHODS}
The inputs of cooperative trajectory prediction, include vector maps and trajectories observed from vehicles and infrastructure, can be represented as $\mathcal{C}=\{\mathcal{T}_V,\mathcal{T}_I,\mathcal{M}\}$. The inputs $\mathcal{T}_V$ are historical trajectories of all objects observed by vehicles over the past $t_h$ time steps: $\mathcal{T}_V=[p^{(1)},p^{(2)},...,p^{(t_h)}]$, where $p^{(t)}=[x^{(t)}_0,y^{(t)}_0,x^{(t)}_1,y^{(t)}_1,...,x^{(t)}_{N_{t}},y^{(t)}_{N_{t}}]$ are the coordinates of all objects at time $t$, and $N_{t}$ is the number of observed objects. Input $\mathcal{T}_I$ and $\mathcal{T}_V$ have the same structure, but the number of objects observed at each time step may differ, so $\mathcal{T}_V,\mathcal{T}_I \in \mathbb{R}^{N_t \times t_h \times C_t}$ while $C_t$ presents the attributes of each trajectory. We represent vector maps by $N_l$ lane segments, the set is denoted as $\mathcal{M} \in \mathbb{R}^{N_l \times 2 \times C_l}$, while $C_l$ represents the attributes of each segment. For multi-modal prediction, the output can be denoted as $Y \in \mathbb{R}^{F,N,t_f,2}$ while $N$ and $F$ represent the number of agents and modes, $t_f$ represents the future time. 
\subsection{Overall Framework}
The proposed framework includes early fusion, PTA-based encoder, mode attention, and anchors decoder. The overview of the framework is illustrated in Fig. \ref{fig:1}(a). Early fusion module receives trajectory inputs from vehicles and infrastructures, along with inputs from vector maps. We match multi-source data based on the distance matrix and use Kalman filtering to fuse the matched data, while unmatched data will also be concatenated to the fused data. Similar to numerous mainstream and exemplary frameworks, our model adopts an encoder-decoder architecture and utilizes graphs to model traffic scenarios. The PTA-based encoder encodes and conducts interactive training on the trajectory data, which is output by the early fusion module and encompasses feature about the agent, lane, and past time. Then we devise a lightweight mode attention module to strengthen the potential connections between different modalities. Finally, a decoder based on sparse oriented anchors is employed to decode the embeddings and output the final multi-agent, multi-modal predicted trajectories. The specific structures of the Anchors decoder and PTA are shown in Fig. \ref{fig:2}(b) and Fig. \ref{fig:2}(c).

\subsection{Early Fusion}
\begin{figure}[t]
	\centering
	{\epsfig{file = 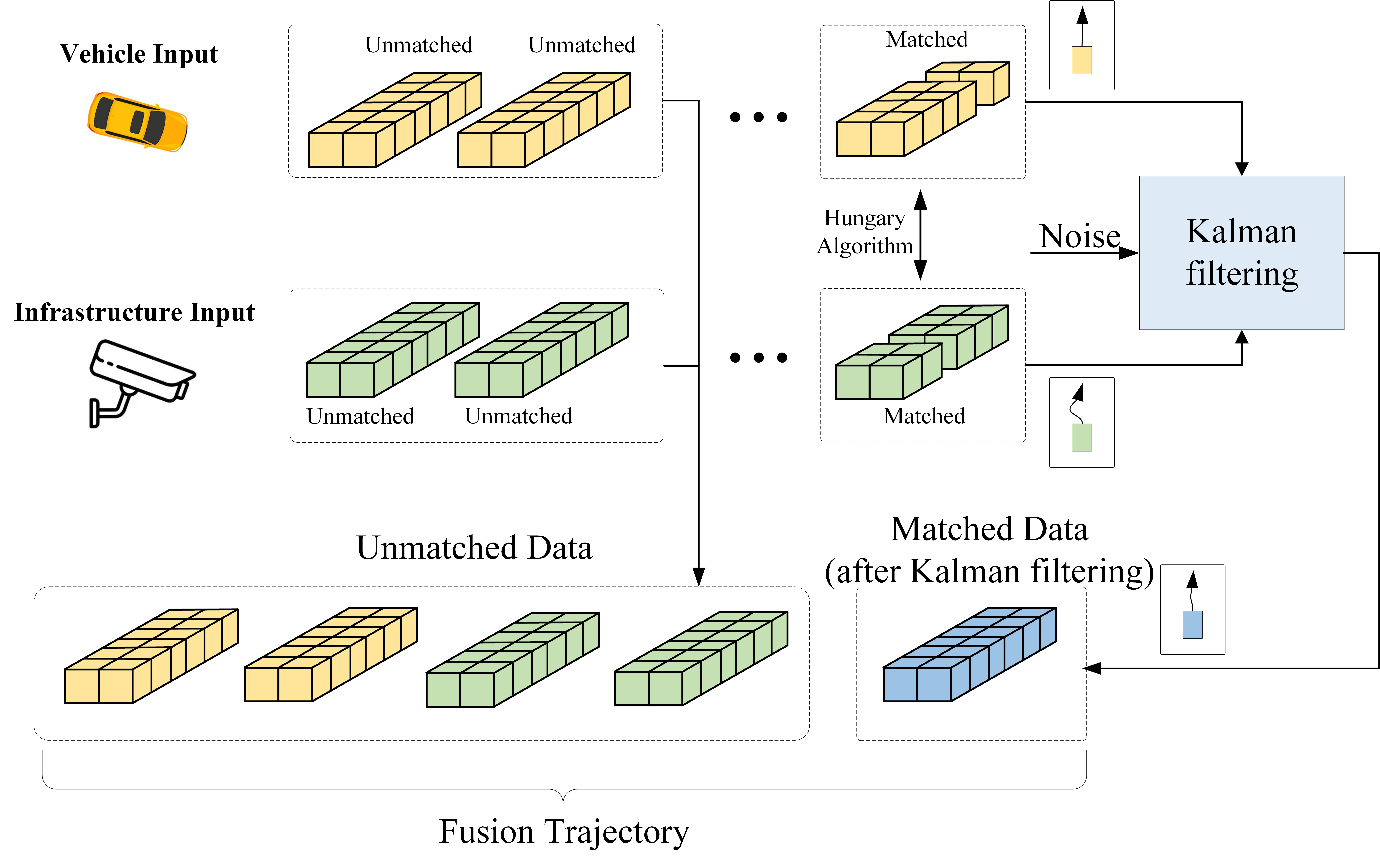, width = 8.5cm}}
	\caption{Overview of early fusion module. Yellow, green, and blue represent vehicle input, infrastructure input and matched data after kalman filtering.}
	\label{fig:3}
\end{figure}

While data captured by infrastructure can effectively complement missing data from vehicles, a common challenge is that the observed trajectories from both sources cannot be directly matched. Currently, a well-established matching method is the Hungarian algorithm, which addresses this particular type of linear programming problem by constructing a cost matrix. However, the extensive amounts of trajectories result in exceedingly high space and time costs when using the Hungarian algorithm. Therefore, we have limited the matching process to the first and last timestamps of the historical trajectories, ensuring that the features of the matched pairs are identical. To achieve more precise fusion results, we fuse the data with Kalman filtering. The overall process can be formulated as:
\begin{equation}\label{eq1}
	\mathcal{T}_{matched}=\mathcal{H}(\mathcal{T}_V,\mathcal{T}_I),
\end{equation}
\begin{equation}\label{eq2}
	\mathcal{T}_{unmatched}=\mathcal{T}_V \cup \mathcal{T}_I-\mathcal{T}_{matched},
\end{equation}
\begin{equation}\label{eq3}
	\mathcal{T}_{fusion}=\mathcal{K}(\mathcal{T}_{matched}) \cup \mathcal{T}_{unmatched},
\end{equation}
where $\mathcal{H}$ and $\mathcal{K}$ represent Hungarian Algorithm and Kalman filtering. The process can be represented as Fig. \ref{fig:3}.

Some works adopt intermediate fusion at the feature level, which results in a multiplication of training parameters \cite{ruan2023learning}. And our experiments reveal that models utilizing early fusion for raw data outperform those that employ intermediate fusion for feature-level data. Furthermore, fusion after trajectory matching can also lead to improved results.

\subsection{PTA-based Encoder}
For each time step in the historical trajectories, a heterogeneous graph is constructed, where each node represents the coordinates of an agent, and the edges connecting nodes denote the relative positional relationships between them. We specify a variable radius, and only edges within this radius will be retained. For the map, we only focus on the roads and intersections where the agent is located, encoding them into a graph using the same method, and including information such as traffic signals in the edges of the graph. The graphs are encoded with multi-layer graph attention networks, aiming to enable the model to comprehensively learn the interaction relationships between agents and maps.

A module called PTA is proposed, aiming to facilitate comprehensive interaction between current and past trajectories in a novel manner. Specifically, the data from all past time steps are first cached. For each time step $t$, the raw data from time steps $t-k_p$ to $t-1$ are concatenated. Additionally, the data at time step $t$ is replicated and expanded to match the same dimension as the concatenated past data. A multi-head cross-attention module is then applied to both to obtain the output of PTA. If time step $t$ is less than $k_p$, the data is directly encoded with an MLP. The process of PTA can be formulated as follows, where $d_t$ represents encoded data at time step $t$ and $\mathcal{R}$ represents repeating to increase dimensionality:
\begin{equation}\label{eq4}
	E_t = \left\{ \begin{array}{l}
		MHCA(\mathcal{C}(d_t,k_p),\mathcal{R}(d_t)),~~~t >= k_p;\\
		MLP(d_t),~~~~~~~~~~~~~~~~~~~~otherwise,
	\end{array} \right.
\end{equation}
\begin{equation}\label{eq5}
	\mathcal{C}(d_t,k_p)=concat(d_{t-k_p},...d_{t-1}),
\end{equation}

\subsection{Mode Attention}
Due to the uncertainty in traffic scenes, each agent may have multiple potential motion directions and speeds in the future, i.e., multiple potential motion modes. With a graph attention module, the data is expanded to multi-modal. As a multi-modal framework, to pay more attention to certain modes, we apply an attention mechanism in the modes. The mode attention mechanism is also represented based on graphs, each trajectory is constructed as a graph. To enhance inference speed, the data is flattened and its dimensions are converted multiple times:
\begin{equation}\label{eq6}
	E_m=GAT(E_t^{(l)},\mathcal{A},\mathcal{D}),
\end{equation}
where $E_t$ represents embeddings output from Encoder, $\mathcal{A}$ and $\mathcal{D}$ are adjacency matrix and attributes on edges, and $GAT$ represents Graph Attention Network.

\subsection{Anchor-oriented Decoder}
Among various trajectory prediction frameworks, anchor-based frameworks \cite{shi2022motion}\cite{zhao2021tnt} have occupied a significant position, providing a novel approach for the final trajectory output method in trajectory prediction, making the output trajectories more realistic and encompassing a more comprehensive set of modes. Anchors are predefined spatial reference points indicating candidate locations for potential future motion trajectories. However, redundant anchors may lead to a decrease in prediction accuracy. Therefore, we set 2 anchors for each mode, which are used to locate the endpoint and midpoint of the future trajectory. We hold the view that a rough prediction can be made based on the prior anchors corresponding to the start, end, and midpoint of a future trajectory, thereby alleviating the burden on the prediction head. Combining the information output by the two above-mentioned modules, we use MLP-Mixer \cite{tolstikhin2021mlp} to complete the multi-modal trajectory prediction for multiple agents.

\subsection{Training and Loss}
Similar to many popular works, the loss of our model during training is composed of multiple components \cite{zhou2022hivt}\cite{ruan2023learning}\cite{tang2024hpnet}\cite{shi2022motion}, as shown in the following equation:
\begin{equation}\label{eq7}
	\mathcal{L}=\mathcal{L}_{cls}+\mathcal{L}_{reg}+\alpha \mathcal{L}_{anchor},
\end{equation}
Firstly, calculate the error between $K$ modes output by the model and ground-truth trajectories, where each mode include $N$ agents and $H$ future time steps, and then select a best prediction mode of shape [N,H,2]. Then the regression loss can be formulated as follows:
\begin{equation}\label{eq8}
	\mathcal{L}_{reg}= - \frac{1}{{NH}}\sum\limits_{k = 1}^K {{\pi_{i,k}}}\sum\limits_{t = {t_h} + 1}^{{t_h} + {t_f}} {\log P(R_i^T(p_i^t - p_i^{{t_h}}|\hat \mu _i^t))},
\end{equation}
where $P(\cdot|\cdot)$ represents the probability density function of the Laplace distribution, $\{\hat \mu _i^t\}^{t_h+t_f}_{t=t_h+1}$ represents the locations of the best mode trajectory for agent $i$, and $R^T_i$ is the rotation matrix \cite{tang2024hpnet}. For classification loss, the probability scores are optimized by using the cross-entropy loss function. Additionally, we use the Huber function to compute the loss at anchors, and multiply it by a coefficient $\alpha$.

\section{\uppercase{Experiments}}
\subsection{Experimental Setup}
\subsubsection{Datasets}
We evaluate cooperative trajectory prediction model over the DAIR-V2X-Seq cooperative trajectory prediction dataset \cite{yu2022dair}, which is the world's first large-scale vehicle-infrastructure cooperative autonomous driving dataset based on real roads. This dataset provides over 60,000 driving scenarios from both vehicles and infrastructures perspectives, with each scene including two intelligent vehicles and one infrastructure perception device. The sampling rate for both perspectives is 10Hz, and the sequence segments are 10 seconds long, providing 5 seconds of historical observations from each view to predict the motion trajectories of all agents in the next 5 seconds. For HD maps, the dataset provides vectorized road information, including road positions, intersection locations, traffic information, etc., and can be accessed using the API provided by Argoverse \cite{chang2019argoverse}.

\begin{figure*}[t]
	\centering
	{\epsfig{file = 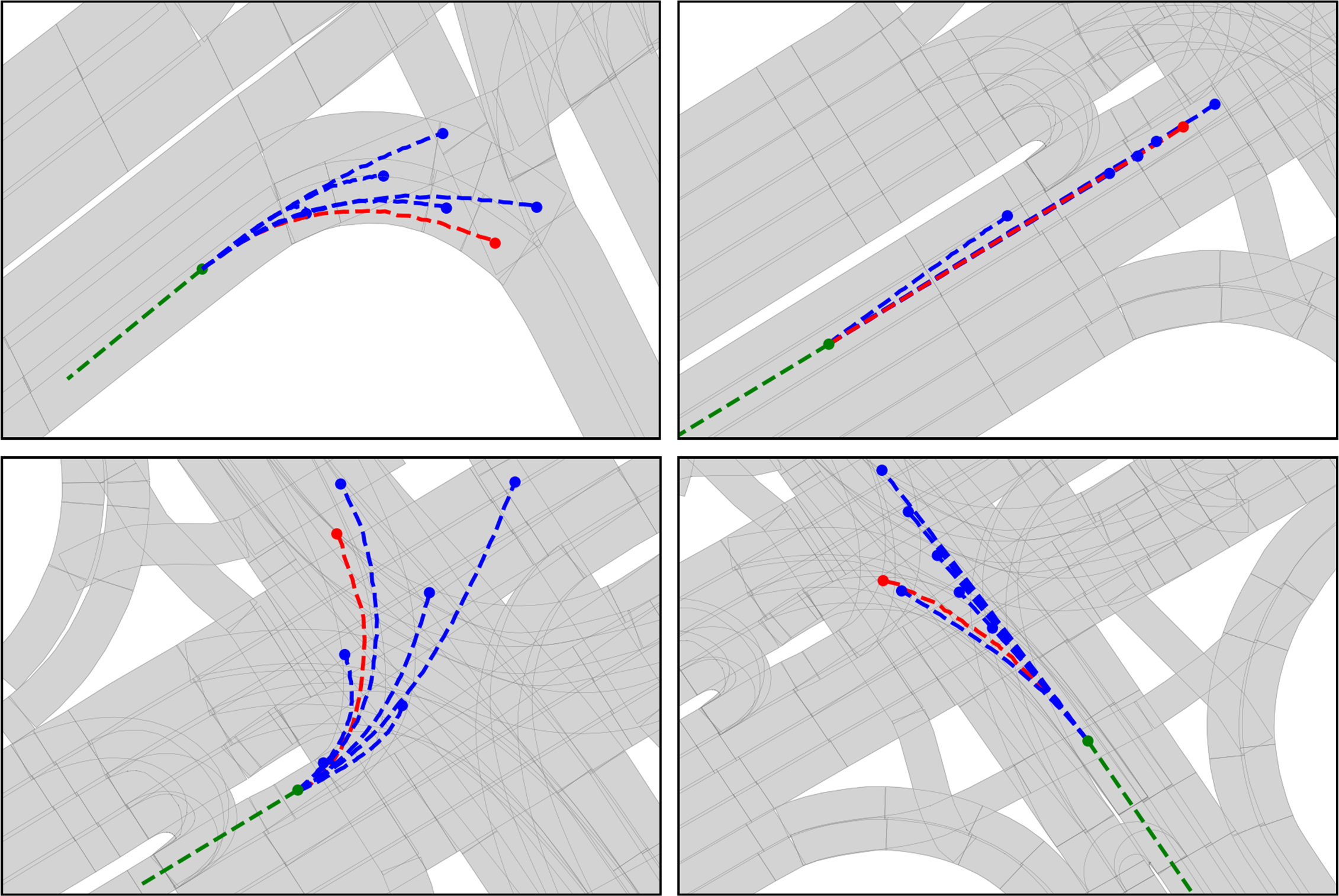, width = \textwidth}}
	\caption{Visualization results over DAIR-V2X-Seq. The past trajectories are shown in green, the ground-truth trajectories are shown in red, and the predicted trajectories are shown in blue.}
	\label{fig:4}
\end{figure*}

\subsubsection{Evaluation Metrics}
The evaluation metrics include Minimum Average Displacement Error (minADE), Minimum Final Displacement Error (minFDE), and Miss Rate (MR). We set 6 final output modes. minADE measures the average ${l}_2$-norm distance between the predicted and actual trajectory points, while minFDE represents the l2 norm distance between the final points of the trajectories. MR evaluates the proportion of cases where the best predicted mode deviates from the actual endpoint by more than 2.0 meters.

\subsection{Main Results}
The model is trained for 64 epochs using AdamW as the optimizer, with a cosine annealing strategy for learning rate scheduling. The initial learning rate is set to 3e-4, the weight decay to 1e-4, and the dropout rate to 0.1. The experimental results in the cooperative scenarios of DAIR-V2X-Seq are shown in Table \ref{table:1}. For comparison, we evaluated classic trajectory prediction models such as TNT \cite{zhao2021tnt}, DenseTNT \cite{gu2021densetnt}, laneGCN \cite{liang2020learning}, HiVT \cite{zhou2022hivt} and the state-of-the-art cooperative prediction model V2X-Graph \cite{ruan2023learning}. To demonstrate the effect of data fusion in V2X scenarios, we evaluated these models using both single-vehicle data and multi-source data, with the early fusion method based on Kalman filtering.
\begin{table}[htbp]
	\centering
	\caption{Quantitative results over DAIR-V2X-Seq}
	\begin{tabular}{p{10.72em}|c|ccc}
		\toprule
		Method & \multicolumn{1}{p{2.4em}|}{Param} & \multicolumn{1}{p{3.3em}}{minADE} &
		\multicolumn{1}{p{3.3em}}{minFDE} &
		 \multicolumn{1}{p{1.7em}}{MR} \\
		\midrule
		TNT (single-vehicle) & \multirow{2}[1]{*}{\textbf{0.5M}} & 8.45  & \multicolumn{1}{c}{17.93} & 0.77 \\
		TNT (cooperative) &       & 7.38  & \multicolumn{1}{c}{15.27} & 0.72 \\
		LaneGCN (single-vehicle) & \multirow{2}[0]{*}{1.0M} & 1.48  & \multicolumn{1}{c}{3.03} & 0.44 \\
		LaneGCN (cooperative) &       & 1.45  & \multicolumn{1}{c}{2.96} & 0.41 \\
		denseTNT (single-vehicle) & \multirow{2}[0]{*}{3.7M} & 1.97  & \multicolumn{1}{c}{3.68} & 0.45 \\
		denseTNT (cooperative) &       & 1.79  & \multicolumn{1}{c}{2.88} & 0.37 \\
		HiVT (single-vehicle) & \multirow{2}[0]{*}{2.6M} & 1.43  & \multicolumn{1}{c}{2.36} & 0.34 \\
		HiVT (cooperative) &       & 1.29  & \multicolumn{1}{c}{2.43} & 0.35 \\
		V2X-Graph (cooperative) & 5.0M   & \textbf{1.17} & \multicolumn{1}{c}{2.03} & \textbf{0.29} \\
		Ours (single-vehicle) & \multirow{2}[1]{*}{3.2M} & 1.42  & 2.30  & 0.33 \\
		\textbf{Ours (cooperative)} &       & 1.24  & \textbf{2.00} & \textbf{0.29} \\
		\bottomrule
	\end{tabular}%
	\label{table:1}%
\end{table}%

Compared to several popular models, in the context of V2X cooperative prediction, our model has achieved the best results on two metrics. It outperforms the state-of-the-art model V2X-Graph on minFDE and achieves comparable results on MR. Additionally, CoPAD is more lightweight with only 64\% of the training parameters compared to V2X-Graph. Compared to HiVT, which ranks third, our model has decreased by 3.9\%, 17.6\%, and 17.1\% on these three metrics, respectively. With the single-vehicle dataset, our model similarly achieves the best results, demonstrating that our model architecture can effectively capture the underlying structure and intent features of the data. Additionally, it can be observed that, compared to using the single-vehicle dataset, all models achieve better performance when using the cooperative prediction dataset. Notably, CoPAD has achieved reductions of 12.7\%, 13.0\%, and 12.1\% in three metrics respectively, which proves that in V2X autonomous driving, the cooperative use of perception results from both vehicles and infrastructures has a positive impact on the prediction model. Specifically, two types of data can complement and correct each other, enabling the deep learning model to mine the historical movement information of the agents more comprehensively.

We visualize the prediction results over DAIR-V2X-Seq, as shown in Fig. \ref{fig:4}. It can be observed that on lanes where both straight and turns are allowed, our model predicts several potentially different trajectories. However, on lanes where only straight travel is permitted, all modes indicate that the vehicle will continue moving straight ahead. Therefore, our model is capable of producing quite diverse future trajectory predictions, which also implies that it can understand and predict the driving intentions of other drivers in the current traffic scenario to a certain extent.

\subsection{Ablation Study}
To further investigate the effectiveness of various modules in our model, we conducted an ablation study, primarily evaluating the contribution of each component in the model, as shown in Table \ref{tab:2}.
\begin{table}[htbp]
	\centering
	\caption{Ablation studies over V2X-Seq. IA and IC respectively represent the intermediate fusion based on data addition and data concatenation. KF represents Kalman filtering and MA means mode attention.}
	\begin{tabular}{p{2.8em}p{1.9em}p{1.7em}c|ccc}
		\toprule
		Fusion & PTA   & MA & \multicolumn{1}{p{3em}|}{Anchors} & \multicolumn{1}{p{3em}}{minADE} & \multicolumn{1}{p{3em}}{minFDE} & \multicolumn{1}{p{1em}}{MR} \\
		\midrule
		None  & \checkmark     & \checkmark     & 2     & 1.42  & 2.31   & 0.33 \\
		IA & \checkmark     & \checkmark     & 2     & 1.49  & 2.43  & 0.35 \\
		IC & \checkmark     & \checkmark     & 2     & 1.48  & 2.42  & 0.35 \\
		KF & \multicolumn{1}{c}{} & \checkmark     & 2     & 1.73  & 3.25  & 0.48 \\
		KF & \checkmark     & \multicolumn{1}{c}{} & 2     & 1.27   & 2.11  & 0.32 \\
		KF & \checkmark     & \checkmark     & 0     & 1.27  & 2.08  & 0.32 \\
		KF & \checkmark     & \checkmark     & 1     & 1.26  & 2.06  & 0.32 \\
		KF & \checkmark     & \checkmark     & 2     & \textbf{1.24} & \textbf{2.00} & \textbf{0.29} \\
		KF & \checkmark     & \checkmark     & 3     & 1.25  & 2.03  & 0.31 \\
		\cmidrule{1-7}    \end{tabular}%
	\label{tab:2}%
\end{table}%

Table \ref{tab:2} demonstrates the superiority of cooperative trajectory prediction once again. We separately encode the multi-source data and then employ addition and concatenation methods respectively to achieve intermediate fusion of the embeddings. Obviously, it is not applicable for scenes characterized by a significant number of missing sampling time steps. In contrast, applying early fusion directly to the raw data provides a more effective means of compensating for missing data. As shown in the table, three components we proposed all improve various metrics to a certain extent, with PTA being particularly effective. After incorporating Mode Attention, the three metrics decrease by 0.03, 0.11 and 0.03 respectively, and PTA further reduces these metrics by 0.49, 1.25, and 0.19 respectively. However, experiments have demonstrated that more anchors does not necessarily mean better performance, setting two anchors at the endpoint and midpoint can achieve optimal results. Compared to three anchors, two anchors result in a reduction of 0.01, 0.03, and 0.02 in minADE, minFDE and MR, respectively. Therefore, ablation studies have demonstrated that our proposed components and fusion paradigm have played significant roles.

\section{\uppercase{Conclusion}}
This paper proposes CoPAD, a lightweight cooperative trajectory prediction framework for V2X cooperative scenes. The Hungarian algorithm and Kalman filtering are employed for data matching and fusion, ensuring an effective integration of information. A PTA-based encoder is designed to augment the interplay between historical data, and a mode attention mechanism is presented to highlight the diversity among multiple modes. Finally, sparse oriented anchors are designed to be integrated with the decoder, facilitating the generation of complete trajectories. The model has achieved impressive performance over the DAIR-V2X-Seq dataset, and future work will delve into the challenges of cooperative prediction in scenes with communication delays.

\section*{ACKNOWLEDGMENT}
This work is supported by National Science and Technology Major
Project, China under Grant 2021ZD0112702 and National Natural Science Foundation (NNSF) of China under Grant 62373100.

\bibliographystyle{IEEEtranS}
\bibliography{ref}

\end{document}